\pdfoutput=1
\documentclass[11pt]{article}

\usepackage[]{ACL2023}

\usepackage{times}
\usepackage{latexsym}

\usepackage[T1]{fontenc}
\usepackage[utf8]{inputenc}

\usepackage{microtype}

\usepackage{amsmath}
\usepackage{mathtools}

\newtheorem{definition}{Definition}

\usepackage{multicol}
\usepackage{subfig}
\usepackage{amssymb}

\title{A Novel Dual of Shannon Information and Weighting Scheme}

 \author {Arthur Jun Zhang, $^{1,2}$\\
\\
\normalsize{${^1}$ Jayoo Technology LLC,}\\
\normalsize{ 21 Fainwood Circle, Cambridge, MA 02139, USA}\\
\normalsize{$^{2}$ Department of Mathematics, Brandeis University,}\\
\normalsize{415 South Street, Waltham, MA 02453, USA}\\
\\
\normalsize{E-mail:  arthurjunzhang@gmail.com.}
}

\begin{document}
\maketitle
\begin{abstract}
Information theory has achieved great success in not only communication technology where it was originally developed for by Shannon but also many other digital fields such as machine learning and artificial intelligence. Inspired by the famous weighting scheme TF-IDF, we discovered that information entropy has a natural dual. We complement the classical Shannon information theory by proposing a novel quantity, namely troenpy. Troenpy  measures the certainty, commonness and similarity of the underlying distribution. To demonstrate its usefulness, we propose a troenpy based weighting scheme for document with class labels, namely positive class frequency (PCF). On a collection of public datasets we show the PCF based weighting scheme outperforms the classical TF-IDF and a popular Optimal Transportation based word moving distance algorithm in a kNN setting. We further developed a new odds-ratio type feature, namely Expected Class Information Bias(ECIB), which can be regarded as the expected odds ratio of the information quantity entropy and troenpy.  In the experiments we observe that including the new ECIB features and simple binary term features in a simple logistic regression model can further significantly improve the performance. The simple new weighting scheme and ECIB features are very effective and can be computed with linear order complexity.  

\end{abstract}

\section{Introduction}
The classical information theory was originally proposed by Shannon\citep{shannon}  to solve the message coding problem in telecommunication. It turned out that it has far more profound impact beyond communication theory, and it has shaped all aspects of our science, engineering and social science by now. The core concept entropy was coined to measure the expected rareness or surpriseness of a random variable $X$ across its distribution. In the literature entropy is usually regarded as the \emph{information}. The mutual information (MI) between two variables is the difference of the entropy of a variable from its conditional entropy given the other variable. MI maximization principle also has been studied and used widely in machine learning. Recently  MI has also been employed as part of the objective function for optimization in neural network models based representation learning\citep{mine,hjelm2018MI}.

Along another line, weighting scheme has been used extensively in information retrieval tasks. Term Frequency-Inverse Document Frequency(TF-IDF), a simple statistic heuristic proposed by \citep{idf1972} has been widely used as a weighting method over half a century in information retrieval and natural language processing. It weighs down a term if its document frequency increases in the corpus, as it becomes less effective to distinguish from others when it gets popular and its appearance brings less \emph{surpriseness} in the sense of Shannon self-information. This simple but effective algorithm has achieved great success as a robust weighting scheme. Even today many search engines and digital database systems still employ TF-IDF as an important default algorithm for ranking.

In the past decades a few researchers have  intensively investigated on it for a better theoretical understanding of the underlying mechanism rather than a heuristic and intuition argument. \cite{rob2004IDF} justified it as an approximate measure of naive Bayes based probability relevance model in information retrieval. Some researchers tried to explain from the information theory point view. \citep{Aizawa2003AnIP} interpreted it as some probability weighted amount of information. \cite{siegler1999} interpreted IDF for a term exactly as the mutual information between a random variable representing  a term sampling and a random variable representing a document sampling from a corpus. Many other variants of the term frequency have been proposed in the literature. For example, BM25\citep{bm25rob} based on probabilistic retrieval framework was further proposed  and it has been widely used by search engines to estimate the relevance of documents to a given search query. In general the derived applications go far beyond text processing and information retrieval community.

The connection between TF-IDF and information theory mentioned above is quite motivating. This makes us wonder if there are other simple and effective weighting schemes that can be established from information theory. In order to achieve this goal, it turns out that we first developed a new concept of information quantity, namely \emph{troenpy}, a natural dual to entropy, and then used it to define a new type of weighting scheme which works very well in the extensive experiments as we hoped.

In the following we will first introduce troenpy, the novel dual of Shannon entropy, and share some insights we have for this innovation. Then for the classical task of supervised document classification, we will develop a troenpy based weighting scheme for document representation. This weighting scheme makes use of the documents class label distribution and helps improving the model performance very significantly. Employing both entropy and troenpy, we will also define some new odds-ratio based class bias features leveraging the document class label distribution. Finally evaluating under the simple KNN and logistic regression settings, we show that  the proposed new weighting scheme and new features are very effective and achieved substantial error reduction compared with the TF-IDF and a popular optimal transportation based document classification algorithm on a collection of widely used benchmark data sets.

\section{Dual of Shannon Entropy}
We fix the notations first. Here we let $X$ indicate a discrete random variables with probability mass function $p_{X}(x)$. The Shannon entropy (sometimes also called self-information) measures the uncertainty of the underlying variable, or the level of  \emph{surpriseness} of an outcome in literature. In
this work we purposely call it Negative Information(NI) for showing the duality nature later. That is, 
\begin{equation}
\textrm{NI}(x):=-log(p_{X}(x))=log\frac{1}{p_X(x)}.
\end{equation}

\noindent Now since Shannon information measures \emph{surpriseness}, can we measure the \emph{commonness} instead? This is exactly the contrary to the Shannon information,  the dual of Negative Information. 
\begin{definition}
We define Positive Information (PI) of an outcome $x$  as  
\begin{equation}
\textrm{PI}(x):=-log(1-p_{X}(x))=log\frac{1}{1-p_{X}(x)}.
\end{equation}
\end{definition}
\noindent So PI has the same value range  $[0, \infty)$ as NI. Note if we denote $\bar{x}$ the complement of outcome $x$, then $\textrm{PI}(x)=\textrm{NI}(\bar{x})$. For discrete random variables with probabilities $p_i$, where $i \in \{1,\dots,K\}$, the value $\frac{1}{1-p_i}$ is the measure of \emph{non-surprisenss} or \emph{commonness}.

Naturally by taking expectation across the distribution,  we propose a dual quantity of entropy, namely \textbf{troenpy}, to measure the certainty of $X$. Troenpy is simply the distributed \emph{positive} information, while entropy measures the distributed \emph{Negative} Information (NI). Troenpy reflects the level of \emph{reliability} of the $X$ outcomes that the data conceals. 

\begin{definition}
The troenpy of a discrete random variable $X$ is defined as the expectation of the PIs,
\begin{equation}
T(x):=-\sum_{x} p_{X}(x)log(1-p_{X}(x)).
\end{equation}
For continuous $X$, the differential troenpy is formally defined correspondingly if the integral is finite,
 \begin{equation}
T(x):=-\int  p_{X}(x) log(1-p_{X}(x))dx.
\end{equation}
\end{definition}

Note conceptually if the certainty increases, it means some outcomes gain more weight and the disorder or uncertainty of the outcomes decreases correspondingly. Because of the intrinsic nature of troenpy, it naturally serves as a weighting scheme measuring the reliability or certainty of a random variable. More certainty means more predictability. If a random variable has very low certainty, this just means it has a high entropy and is very noisy.   Thus  it is not a good candidate for prediction purposes and should be correspondingly down-weighted. 

\begin{figure*}[ht]
\centering
\includegraphics[width=1.0\linewidth]{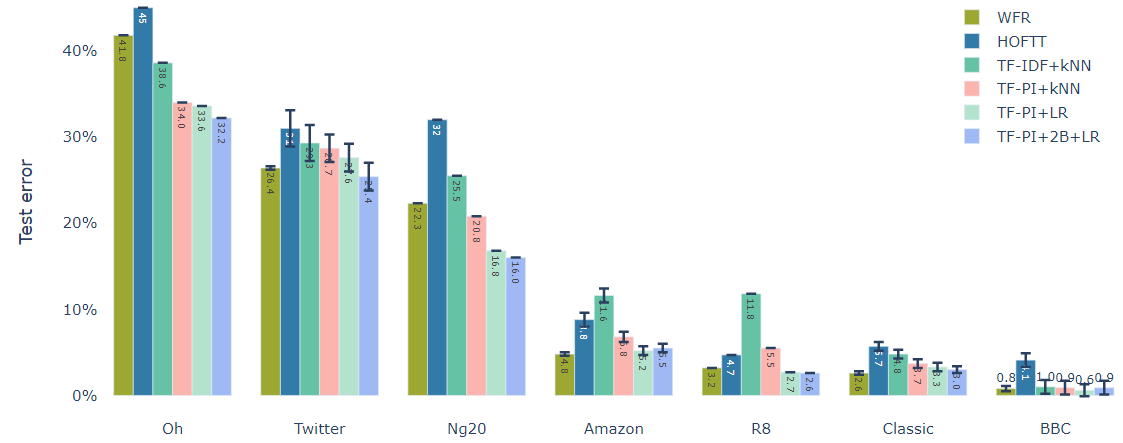}
\caption{Errors of document classification for 7 Datasets with TF-IDF and TF-PI}
\end{figure*}

\section{Weighting Scheme for Supervised Documents Classification }
In this section we first briefly review the information theoretic interpretation of TF-IDF, then naturally we define a new weighting scheme using the newly proposed troenpy as an analogue.
\subsection{Review of IDF}
Here we follow the information theoretic view mentioned above \citep{Aizawa2003AnIP}. We consider the classical text documents classification task in the routine supervised learning setting.  The typical scenario is that given a corpus collection of documents $\text{D}_1, \dots,\text{D}_n$, where $\text{n}$ denotes the total number of documents. Each document $\text{D}_i$ has a class label $y_i$ from a finite class label set $Y=\{1,2,\dots,K\}$, where $K$ is the total number of classes. For a given word term w, let $d$ denote the number of documents where $w$ appears. Then the IDF is simply given by the following:
\begin{equation}
IDF(w)=1+log\frac{n}{1+d}
\end{equation}
 
 It can be interpreted as the \emph{self}(negative) information in information theory, which measures the \emph{surprisal} of the term t. The idea follows as below: Fix a word $w$ with document frequency d in a collection of $N$ documents, then the probability of $w$ appear in a document D can be approximated by $Prob(w \in D)=\frac{d}{N}$. Then the negative-information $NI(w)=-logProb(w\in D)=log\frac{N}{d}$. To smooth out the case when $d=0$, adding 1 to the denominator gives $NI(x)\approx log\frac{N}{d+1}$. Also, the summation of all TF-IDFs, each of which represents bits of information weighted by the probability of a term, also recovers the mutual information between terms and documents.

\subsection{Positive Class Frequency}
In this section we will make use of the document class distribution and define a new term weighting method, which can be applied later for the classification task. First for all the $n$ documents in the corpus, we collect the counts of documents for each class. We denote the class label distribution  as $C=\{C_1,\dots,C_K\}$, where $C_i$ is the count of the $i^{th}$ class label. Normalizing by dividing the total number of documents $n$ gives the probability distribution $\overrightarrow{c}=\{c_1,\dots,c_K\}$, where $c_i=\frac{C_i}{n}$. This vector $\overrightarrow{c}$ contains the global distribution information and we can define two intrinsic quantities measuring the certainty and uncertainty.
\begin{definition}
We define Positive Class Frequency(PCF) for $C$ as the troenpy of $\overrightarrow{c}$. Similarly, Negative (or Inverse) Class Frequency(NCF or ICF) as the entropy of $\overrightarrow{c}$.  
\begin{equation}
\begin{split}
\textrm{PCF}(C):&=Troenpy(c)\\
\textrm{NCF}(C):&=Entropy(c)\\
\end{split}
\end{equation}
\end{definition}

For the whole documents collection (abbreviated as $\textrm{DC}_*$), the PCF of the normalized label vector $\overrightarrow{c}$, denoted as $PCF_*$ is a constant for each term indicating the certainty level of the whole label distribution at the collection population level. Restricting to the documents with the term $w$ present (abbreviated as $DC_1$), the corresponding PCF is denoted as $\textrm{PCF}_1$. Similarly, $\textrm{PCF}_{-1}$ denotes the PCF for documents without the term $w$ (abbreviated as $DC_{-1}$). We propose using the difference $\textrm{PCF}_1-\textrm{PCF}_*$ between $\textrm{PCF}_1$ and $\textrm{PCF}_*$ as a term weighting reflecting the certainty gain due to the presence of the term $w$. Without abuse of notation, we simply keep using PCF to denote this new weighting scheme. It can also be interpreted as a conditional troenpy condition on the knowledge of the presence or absence of the term $w$. Note in the classical TF-IDF setting and general machine learning literature, such label distribution information has never been made use of before. 

To combine the IDF and PCF weightings, we propose using their multiplication $\text{PCF}\cdot\text{IDF}$, abbreviated as \textbf{PI}, as the weighting. Hence multiplying with the term frequency gives the name \textbf{TF-PI}. So in our setting each document can be represented as a vector of word term frequencies multiplied with selected weighting method applied such as $doc_i=[tf_1\textrm{PI}_1,\dots,tf_m\textrm{PI}_m]$, where $m$ is the number of unique selected terms in a document. 

On the other hand, the entropy based NCF is not suitable for weighting as they are the negative information measuring the uncertainty. We will illustrate and explain this phenomenon elsewhere \cite{ITC}.

\section{Class Information Bias Features and Binary Term Frequency Features}
In this section we introduce two types of features for document representation: the  odds ratio based features for class information distribution and a simple binary term frequency feature. For abbreviation, we denote these two features as 2B features in the experiments.

\subsection{Odds-Ratio based Class Information Bias Features}
The idea is that both the TF-IDF and TF-PI are obtained from a term frequency multiplied with a weight information quantity measuring their rareness or certainty, instead we can weight these term frequencies by how biased they distributed across the classes. 
This idea was inspired by an algorithm called Delta-IDF. In a simple two class sentiment classification setting, \cite{deltaidf} proposed first taking the difference of the IDFs between the documents of the positive class and the documents of the negative class and then multiplying with the term frequency to give their delta-TFIDF. That is, $tf_w[log\frac{P}{P_w}-log\frac{N}{N_w}]$, where $P$ and $N$ respectively stand for the total numbers of positive documents and negative documents, and the $P_w$ and $N_w$ respectively stand for the total numbers of positive documents with the term $w$ appears and the total number of negative documents with term $w$ appears. So the difference between the IDFs of the two collections of documents are exactly the odds ratio of the documents counts for the two complementary collections of documents, which can be rewritten as $log\frac{PN_w}{P_wN}$. 

Motivated by the above, we  can first compute the NCF and PCF difference for any class $i$, which gives the the Class Information Bias (CIB) features. And then we take the weighted average of such CIB features across all $K$ classes. We call these new features the Expected Class Information Bias (ECIB) features. Specifically for a term $w$, we first use $n_w$ denote the number of documents with $w$ present and $n_{iw}$ denote the number of documents with class label $i$ and w present. Then the NCF based CIB for class $i$ is given as $\textrm{CIB}_i(w)=log\frac{C_i}{1+n_{iw}}-log\frac{n-C_i}{1+n_w-n_{iw}}$, as $(n-C_i)$ stands for the total documents not in class $i$ and $(n_w-n_{iw})$ stands for the total number of documents not in class $i$ but with $w$ appears. Similarly, the PCF based CIB is given as $log\frac{C_i}{1+C_i-n_{iw}}-log\frac{n-C_i}{1+n-C_i-n_w+n_{iw}}$.

Therefore for each term $w$, we can define two such distributed Class Information Bias features, one using NCF and one using PCF. The expected CIB features are precisely given by the following.

\begin{equation}
\begin{split}
& \text{CIB-NCF}(w):\\
&=\sum^{K}_{i=1}\frac{C_i}{n}(log\frac{C_i}{1+n_{iw}}-log\frac{n-C_i}{1+n_w-n_{iw}})\\
& \text{CIB-PCF}(w):=\sum^{K}_{i=1}\frac{C_i}{n}(log\frac{C_i}{1+C_i-n_{iw}}\\
&  -log\frac{n-C_i}{1+n-C_i-n_w+n_{iw}})\\
\end{split}
\end{equation}

The effect for this ECIB feature is that words that are evenly distributed for their contribution of the information quantities in a class and the rest of the class  get little weight, while words that are prominent in some class for their contribution of the information quantities get more weight. So the terms characterizing specific classes are relatively better weighted as they are more representative.

\subsection{Binary Term Frequency}
The binary term frequency (BTF) is simply a binary feature for each term $w$. BTF($w$) is 1 if w is present in a document and it is 0 if it is absent in a document. BTF gives the most naive representation of a document, regardless of frequency counts. We notice that BTF features are actually quite informative and together with TF-IDF can significantly improve the classification performance in the kNN setting. One can achieve this by simply summing the TF-IDF based pairwise document distance and the BTF features based document pairwise distance as the final document pairwise distance. 

\begin{figure*}[ht]
\centering
\includegraphics[width=1.0\linewidth]{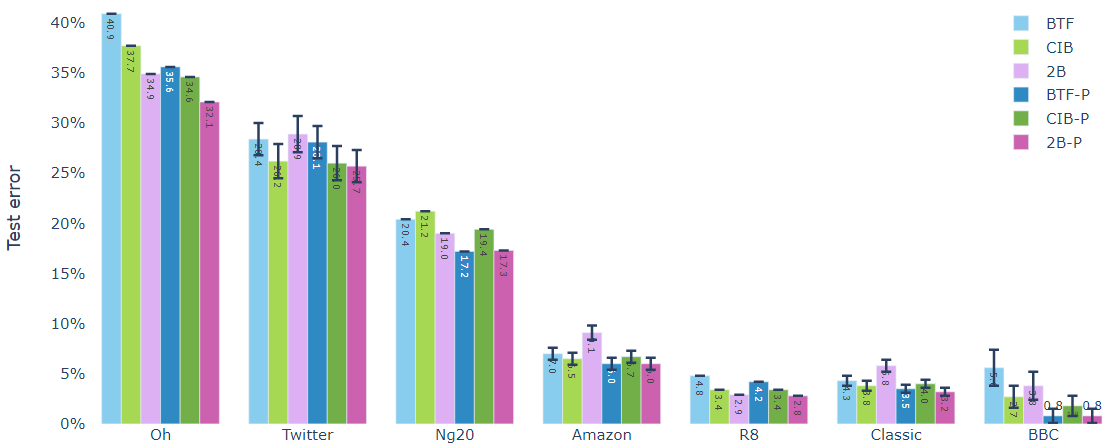}
\caption{Error rates of document classification using 2B features in logistic regression classifier}
\end{figure*}

\section{Datasets and Experiment}
The goal of our experiments in this section is to validate our proposed weighting schemes and features for the supervised document classification tasks,  and compare with the baseline algorithms.   To achieve this we include seven text document datasets that are often used for the documents classification tasks in the literature. Three datasets already have a training dataset and a test dataset split while the rest four have no such splits. The experiments of supervised document classification tasks have two settings for the evaluation:  a simple kNN setting and a logistic regression setting. The evaluation metric is the error rates on the test datasets.

\subsection{Datasets}
Here we follow closely the setup of \cite{Mikhail2019}.
We use the popular seven open source datasets below for the study on KNN based classification tasks. Note these datasets have been extensively used numerous times for the classification tasks. The datasets include BBC sports news articles labeled into five sports categories (BBCsports);  medical documents labeled into 10 classes of cardiovascular disease types( Ohsumed); Amazon reviews labeled by three categories of Positive, Neutral and Negative (Amazon);  tweets labeled by sentiment categories (Twitter);  newsgroup articles labeled into 20 categories (20 News group);  sentences from science articles labeled by different publishers ( Classic) and Reuters news articles labeled by eight different topics (R8). The more detailed information about the datasets can be  found in the references mentioned above. For the datasets with no explicit train and test splits, we use the common 80/20 train-test split and report the performance result based on fifty repeats of random sampling.

To minimize the datasets version mismatch, in all the experiments we use the raw text documents rather than some pre-processed intermediate formats such as some of the processed datasets provided in \cite{kusner2015doc}.

\begin{figure*}[ht]
\centering
\includegraphics[width=7.5cm]{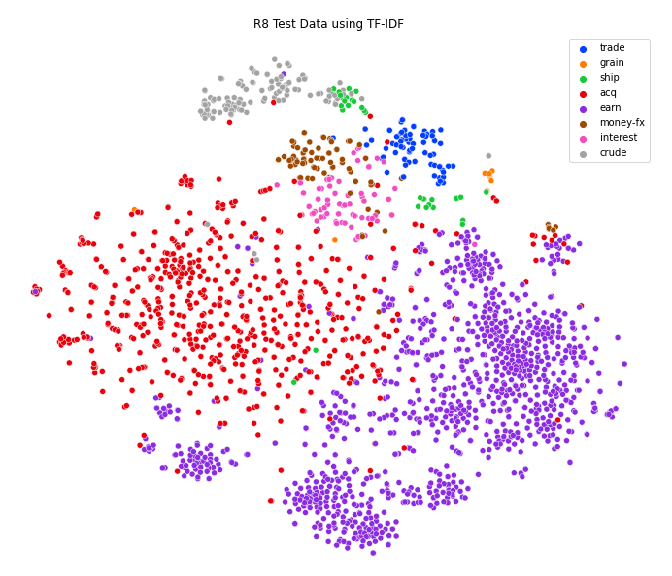}
\qquad
\includegraphics[width=7.5cm]{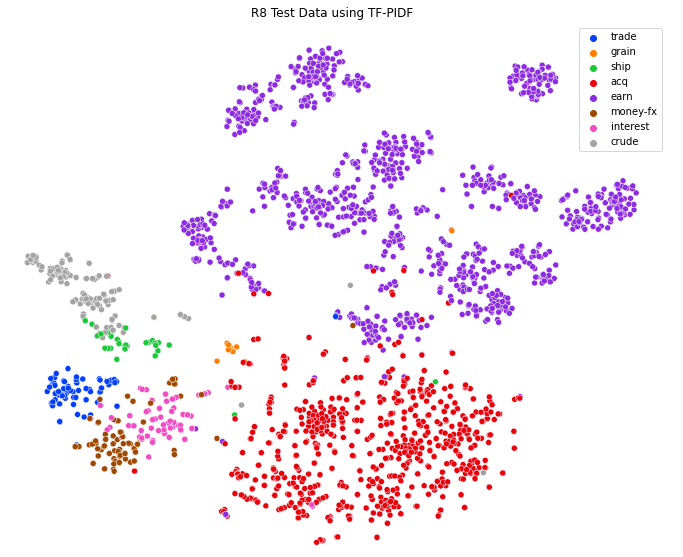}
\caption{ t-SNE on R8 data}
\end{figure*}

\subsection{Experiment Settings}
For the evaluation of supervised documents classification on term frequencies and their weighting, we include the classical TF-IDF document representation as a baseline. The pairwise document distance in kNN setting is computed using the TF-IDF represented vectors. For comparison purpose and reference, in the experiments we also include the result of a Word Moving Distance (WMD) based algorithm, namely HOFTT proposed by \cite{Mikhail2019}. It is a hierarchical optimal transportation distance in the topic spaces of documents. The experiment setting of HOFTT is identical to ours.

\paragraph{kNN based Classification:}
The features include term frequencies only. The goal is to validate the TF-PI weighting and compare with TF-IDF.
The data pre-processing starts with removing the frequent English words in the stop word list, which can be found in the above references. To ease the kNN evaluation part, we fix the number of closest neighborhoods K=7 rather than dynamically selecting the optimal K. We compute the integrated weighting PI as the product of PCF and IDF, and compare with the IDF weighting for each term frequency. Using the TF-PI and TF-IDF, we obtain the bag-of-words vector representation of each document and take their L2 normalization, and then  compute the document pairwise distance following the standard kNN procedures. Again our main goal here is to assess if the proposed PCF weighting is effective and can help improve the classical TF-IDF method.

\paragraph{Logistic Regression based Classification:} 
In this setting we simply replace the simple kNN with a standard logistic regression model instead. Here we have two goals to evaluate. First we need to evaluate if the models have performance improvement when the 2B features are included, compared with the models using only the TF-PI features. So we can assess if 2B features are effective for the document classification task. Second we want to evaluate the PCF weighting effect on the ECIB and BTF features both separately and jointly. 

Here the data preprocessing is identical to the kNN classification settings above.  We mainly consider three types of features in the experiment, namely the TF-PI features, binary term features (BTF) and the ECIB features.

\section{Results and Discussion}

\paragraph{kNN based Classification Experiments:}
In Figure 1 we can visually observe that  the TF-PI based kNN model uniformly outperformed the classical TF-IDF based kNN across all seven datasets and the improvement is quite substantial for most cases with an average overall error reduction $22.9\%$. Noticeably the R8 dataset achieves the most $53.4\%$ error reduction. These uniform improvement can be explained as the PCF weighting does effectively leverage the certainty and common similarity of class label distributions across the corpus at a term level. For a term, the more PCF it has the better prediction capacity it has. For example, the different news groups in Ng20 actually share many non-stop words in common and some groups are very relevant. The learned similarity information about one group is helpful at predicting a relevant group. We also observe only slight improvement on the Twitter and BBC sport datasets which might be simply due to the small sample sizes. The Twitter has 3115 samples and BBCsport has only 737 samples, which are quite small compared with other datasets. Additionally, the Twitter sentiment dataset has three class labels consisting of positive, neutral and negative. The extreme polarity of the classes is often consistent with the fact that relatively less common description words are shared across the classes. 

\paragraph{t-SNE:} We also use the popular t-SNE by \cite{tsne2018} to visualize the TF-IDF and TF-PI classification effect on the R8 dataset. In Figure 3, the TF-PI appears to cluster relatively closer for each class labels and clusters are relatively separated from other cluster groups. 

\paragraph{Word Moving Distance Methods:} In the experiments a modern Optimal Transportation (OT) based Word Moving Distance (WMD) approach HOFTT performs poorly compared with the TF-PI weighting on all dataset except on R8 dataset, on which it is also outperformed by TF-PI employing the additional 2B features. However we are also aware another advanced WMD method Wasserstein-Fisher-Rao(WFR) developed by \citet{wfr8zihao}, which uses the Fisher-Rao metric for the unbalanced optimal transportation problem. The reported result of WFR is comparable to our proposed methods across the datasets. Unfortunately there are some version mismatch for some datasets as well as slightly different sampling procedure for datasets with no pre-specified train-test splits, so we did not include the corresponding result in our figures. Note also that the general Sinkhorn based algorithms for such OT optimization problems have relatively high computational complexity and so they are quite expensive on computational cost. While the proposed weighting scheme and ECIB features can be obtained in a single scan of the data and the time complexity is linear, they are fast and  a lot cheaper on computational cost.

\paragraph{Logistic Regression based Experiments:}
In Figure 1 we observed the following:  (1) for the same TF-PI feature set, the logistic regression model uniformly outperforms the kNN approach across all datasets. This is not surprised as the logistic regression optimizes the term coefficients for optimal fitting the data while the kNN is rigid as given. (2) adding the 2B features of binary term frequency (BTF) and expected class information bias (ECIB) further significantly reduces the errors on most datasets. For the BBC dataset we observed a relatively large error increase, and we hypothesize that this may be due to the very small test sample size of the dataset.   

In Figure 2 we reported the results of using BTF and ECIB features in the logistic regression setting. We observed the following. Both BTF and ECIB features are quite effective when used individually alone. ECIB performs better than BTF on all datasets except on the dataset of 20 Newsgroup, where they are relatively close. Simply combining the two features together not necessarily always improves the performance, instead it leads to slightly more errors on  a couple of the datasets. We also observe that applying the PCF weighting helps on majority of the cases. Visually the left three bars of light color represent 2B features without PCF weighting while the right three bars of darker color represent corresponding features with PCF weighting applied.

\section{Conclusion}
We proposed the new concept troenpy as the dual of the Shannon entropy for a distribution, which summarizes the certainty or common similarities of the outcome distribution. We also developed a troenpy based simple weighting scheme PCF for documents with labels, which can leverage the document class distribution information across a corpus. Combining the PCF and the well-known TF-IDF gives a new  term frequency based document representation. For the document classification task, this TF-PI weighting method significantly outperformed not only the classical TF-IDF weighting method in a fixed KNN setting but also an advanced optimal transportation based word moving distance method on a  collection of widely used datasets. 

Additionally we also proposed two new features: binary term presence(BTF) feature and expected class information bias (ECIB) feature. The ECIB features are the weighted odds-ratios of the information quantities across all classes, so they weigh the terms according to how much distance they are from the evenly contribution to the two information measures. In a simple logistic regression setting we demonstrated that both features are quite effective when used alone for the document classification task while their combination together with the TF-PI features performs the best on the datasets with further significant error reduction. The proposed algorithms have only linear complexity and run fast. They can be applied to extract useful information from the data with only single scan of the data and then be easily integrated as weightings into many other machine learning tasks.

\section*{Limitations}
The current work proposed a new weighting scheme leveraging the document label information and demonstrated the effectiveness on a popular benchmark collection of English datasets. For documents without label information available, the current proposal cannot apply directly. However, these unsupervised tasks often can be reformulated into popular self-supervised problems. And then we can  apply the developed methods to process. A complete demonstration and explanation is treated elsewhere \cite{ITC}.

For image processing with pixels values in the typical range [0,255] or other continuous data features, straight applying the above weighting schemes does not work. More methods are needed to be developed for problems in such scenarios. 
It will be interesting to check the performance of the proposed algorithms on some other large scale corpus of data-sets  and  also data-sets in other languages. The weighing scheme application are mainly dealt with the discrete tokens scenarios and it is not clear how these  methods can extend to the modern distributed representation scenarios where each token is encoded as a high dimensional vector or tensor in general. Especially the pretrained word vectors or transformer based models are dominating the NLP community, it is not clear for the weighting methods how to integrate with these trends.

\section*{Ethics Statement}
The conducted research tries to find principled ways for natural language computing tasks for the general goods  and technology advancement of the society. 

\section*{Acknowledgments}
AJZ thanks Timothy J. Hazen for first introducing TF-IDF to him in 2013. AJZ thanks an anonymous reviewer for several tips on improving the manuscript presentation as well as the inspiring question of how applying troenpy to unlabeled data. The related USPTO Patent 18161067 is pending.

\bibliography{anthology,mybib}
\bibliographystyle{acl_natbib}
\appendix

\end{document}